\documentclass[conference]{IEEEtran}
\usepackage{times}
\usepackage{color, colortbl, xcolor}

\usepackage[numbers]{natbib}
\usepackage{multicol}
\usepackage[bookmarks=true]{hyperref}
\usepackage{amsfonts}
\usepackage{amsmath}
\usepackage{dsfont}
\usepackage{subfigure}
\usepackage{caption}
\usepackage{multicol}
\usepackage[pdftex]{graphicx}   
\usepackage{graphicx}

\graphicspath{{./figs/}}    
\DeclareGraphicsExtensions{.pdf,.png}

\usepackage[textfont=md,font=footnotesize]{caption}

\usepackage{ifthen}
\newboolean{include-notes}
\setboolean{include-notes}{true}

\newcommand{\adnote}[1]{\ifthenelse{\boolean{include-notes}}{\textcolor{purple}{\textbf{A:#1}}}{}}

\usepackage{lipsum}

\newcommand\blfootnote[1]{%
  \begingroup
  \renewcommand\thefootnote{}\footnote{#1}%
  \addtocounter{footnote}{-1}%
  \endgroup
}

\newcommand{\dfknote}[1]%
    {\textcolor{green}{\textbf{DFK: #1}}}
\newcommand{\abnote}[1]%
    {\textcolor{cyan}{\textbf{AB: #1}}}
\newcommand{\jfnote}[1]%
    {\textcolor{blue}{\textbf{JF: #1}}}
\newcommand{\shnote}[1]%
    {\textcolor{orange}{\textbf{SH: #1}}}
\newcommand{\remove}[1]%
    {\textcolor{red}{#1}}

\newcommand{\example}[1]%
{
\textbf{Running example:}
\textit{#1}
}

\begin{document}


\title{Probabilistically Safe Robot Planning\\ with Confidence-Based Human Predictions}

\author{

\authorblockN{
Jaime F. Fisac\authorrefmark{1} \qquad
Andrea Bajcsy\authorrefmark{1}\qquad
Sylvia L. Herbert\qquad
David Fridovich-Keil \\
Steven Wang \qquad
Claire J. Tomlin \qquad
Anca D. Dragan
}
\authorblockA{Department of Electrical Engineering and Computer Sciences\\
University of California, Berkeley\\
\{
\href{mailto:jfisac@berkeley.edu}{jfisac},
\href{mailto:abajcsy@berkeley.edu}{abajcsy},
\href{mailto:sylvia.herbert@berkeley.edu}{sylvia.herbert},
\href{mailto:dfk@berkeley.edu}{dfk},
\href{mailto:sh.wang@berkeley.edu}{sh.wang},
\href{mailto:tomlin@berkeley.edu}{tomlin},
\href{mailto:anca@berkeley.edu}{anca}
\}@berkeley.edu}
}


%

\maketitle



\begin{abstract}
In order to safely operate around humans, robots can employ predictive models of human motion.
Unfortunately, these models
cannot capture the full complexity of human behavior and necessarily introduce simplifying assumptions.
As a result, predictions may degrade whenever the observed human behavior departs from the assumed structure,
which can have negative implications for safety.
In this paper, we observe that
how ``rational'' human actions appear under a particular model
can be viewed as an indicator of that model's ability to describe the human's current motion.
By reasoning about this \emph{model confidence} in a real-time Bayesian framework, we show that the robot can very quickly modulate its predictions to become more uncertain when the model performs poorly.
Building on recent work in provably-safe trajectory planning, we leverage these confidence-aware human motion predictions to generate assured
autonomous robot motion.
Our new analysis combines worst-case tracking error guarantees for the physical robot with probabilistic time-varying human predictions, yielding a quantitative, probabilistic safety certificate. We demonstrate our approach with a quadcopter navigating around a human.
\vspace{-.5cm}
\end{abstract}

\IEEEpeerreviewmaketitle



\blfootnote{This research is supported by an NSF CAREER award, the Air Force Office of Scientific Research (AFOSR), NSF's CPS FORCES and VehiCal projects, the UC-Philippine-California Advanced Research Institute, the ONR MURI Embedded Humans, and the SRC CONIX Center.}
\blfootnote{
    $^*$The first two authors contributed equally to this paper.
}

\section{Introduction}
\label{sec:intro}

\begin{figure}[t!]
    \centering
    \includegraphics[width=\columnwidth]{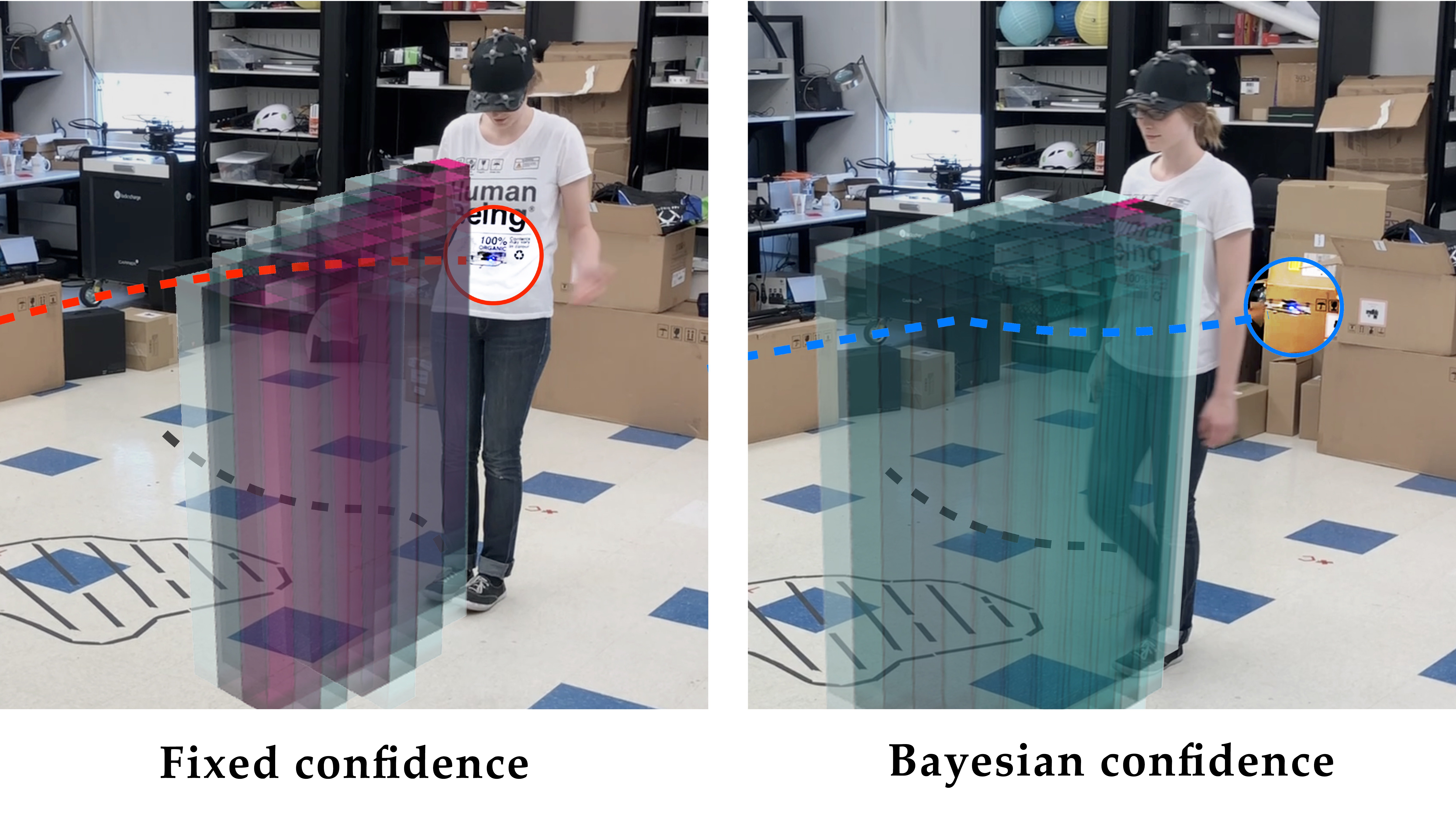}
    \caption{When planning around humans, predictive models can enable robots to reason about future motions the human might take. These predictions rely on human motion models, but such models will often be incomplete and lead to inaccurate predictions and even collisions (left). Our method addresses this by updating its \textit{human model confidence} in real time (right).}
    \label{fig:front_fig}
    \vspace{-.7cm}
\end{figure}

In situations where robots are operating in close physical proximity with humans, it is often critical for the robot to anticipate human motion.
One popular predictive approach is to model humans as approximately rational with respect to an objective function learned from prior data \cite{ ziebart2009planning, kretzschmar2016socially}.
When a person is moving in accordance with the learned objective (e.g. to a known goal location), such models often make accurate predictions and the robot can easily find a safe path around the person.
Unfortunately, no model is ever perfect, and the robot's model of the human will not be able to capture all possible movements that it might eventually observe.
For example, the human might walk toward another goal location that the robot does not know about, or move to avoid an obstacle of which the robot is unaware. In these cases where the human's motion diverges from the model's predictions, safety might be compromised. In Fig.~\ref{fig:front_fig} (left), the robot fails to reason about the human avoiding the unobserved obstacle and gets dangerously close to the human.

One method to mitigate the effects of model inaccuracy is for the robot to re-compute its human model over time. However, restrictions in sensing and in the availability of human data limit how much a model can be refined online without overfitting. 
Alternatively, the robot can reason about its confidence in its current model's predictions.
In this paper, we propose a method in which the robot continually estimates its confidence in its human model in real time and adapts its motion plan according to this confidence (Fig.~\ref{fig:front_fig}, right).
In particular, our approach leverages the so-called ``rationality'' coefficient in the commonly used Boltzmann model of approximately rational human behavior \cite{baker2007goal,ziebart2008maximum}
%
as a time-varying indicator of the model's predictive performance.
This is a single scalar parameter that can
be tractably inferred at deployment time.
We couple the resulting confidence-aware human motion predictions with a provably safe motion planner to obtain \textit{probabilistically safe} robotic motion plans that are conservative when appropriate but efficient when possible.

\textit{This paper makes two key contributions:}
(1) a real-time Bayesian framework for reasoning about the uncertainty inherent in a model's prediction of human movement,
and
(2)~extending a state-of-the-art, provably safe, real-time robotic motion planner to incorporate our time-varying, probabilistic human predictions.
Together, these two contributions facilitate the real-time generation of robot trajectories through human-occupied spaces.
Further, they guarantee that when the robot tracks these trajectories at run-time they will be collision-free with arbitrarily high probability.

\section{Prior Work}
\label{sec:related}

\subsection{Human Modeling and Prediction}

One common approach for predicting human actions is supervised learning, where the current state and the history of the human's actions are used directly to predict future actions. Such approaches have enabled inference and planning around human arm motion \cite{amor2014interaction,ding2011human,koppula2013anticipating,lasota2015analyzing,hawkins2013probabilistic},  navigation \cite{ding2011human}, plans for multi-step tasks like assembly \cite{hawkins2013probabilistic}, and  driving \cite{Schmerling2017}.

Rather than predicting actions directly, an alternative is for the robot to model the human as a rational agent seeking to maximize an unknown objective function. The human's actions up to a particular time may be viewed as evidence about this objective from which the robot may infer the parameters of that objective. Assuming that the human seeks to maximize this objective in the future, the robot can predict her future movements \cite{baker2007goal, ng2000algorithms}. 
In this paper, we build on in this work by introducing a principled online technique for estimating confidence in such a learned model of human motion.

\subsection{Safe Robot Motion Planning}
Once armed with a predictive model of the human motion, the robot may leverage motion planning methods that plan around uncertain moving obstacles and generate real-time dynamically feasible and safe trajectories.

To avoid moving obstacles in real time, robots typically employ reactive and/or path-based methods. Reactive methods directly map sensor readings into control, with no memory involved \cite{belkhouche2009reactive}. Path-based methods such as rapidly-exploring random trees and A* find simple kinematic paths through space and, if necessary, time \cite{hart1968astar,karaman2011RRTPRM}.
These path-based methods of planning are advantageous in terms of efficiency,
yet, while they have in some cases been combined with probabilistically moving obstacles \cite{Aoude2013probabilistically,ziebart2009planning},
they do not consider the endogenous dynamics of the robot or exogenous disturbances such as wind. As a result, the robot may deviate from the planned path and potentially collide with obstacles. It is common for these plans to try to avoid obstacles by a heuristic margin of error. FaSTrack is a recent algorithm that provides a guaranteed tracking error margin and corresponding error-feedback controller for dynamic systems tracking a generic planner in the presence of bounded external disturbance \cite{herbert2017fastrack, fridovich2018planning}. Our work builds upon FaSTrack to create an algorithm that can safely and dynamically navigate around uncertain moving obstacles in real time.

\section{Problem Statement and Approach}
\label{sec:problem}
We consider a single robot moving to a preset goal location in a space shared with a single human, and assume that the human expects the robot to avoid her.
Therefore, it is the robot's responsibility to maintain a safe distance from the human at all times.
We present our theory for a general single human and single robot setting, and use the running example of quadcopter navigating around a walking human to illustrate the proposed approach and demonstrate the utility of our method.

\subsection{Motion Model}
\label{subsec:robot_motion_model}

Let the state of the human be $x_H \in \mathbb{R}^{n_H}$, where $n_H$ is the dimension of the human state space. 
We similarly define the robot's state, for planning purposes, as  $x_R \in \mathbb{R}^{n_R}$.
These states could represent the positions and velocities of a mobile robot and a human in a shared environment or the kinematic configurations of a human and a robotic manipulator in a common workspace.
The human and robot are each modeled by their dynamics:
\begin{equation}
    \begin{aligned}
    \dot{x}_H &= f_H(x_H, u_H) \qquad
    \dot{x}_R &= f_R(x_R, u_R)
    \end{aligned}
\end{equation}
where $u_H \in \mathbb{R}^{m_H}$ and $u_R \in \mathbb{R}^{m_R}$ are the control actions of the human and robot, respectively.

The robot ultimately needs to plan and execute a trajectory to a goal state
according to some notion of efficiency,
while avoiding collisions with the human. 
We define the keep-out set $\mathcal{K} \subset \mathbb{R}^{n_H} \times \mathbb{R}^{n_R}$ as the set of joint robot-human states to be avoided, e.g. because they imply physical collisions.
To avoid reaching this set, the robot must reason about the human's future motion when constructing its own motion plan.

\example{
We introduce a running example for illustration throughout the paper.
In this example we consider a small quadcopter that needs to fly to locations $g_R \in \mathbb{R}^3$ in a room where a human is walking.
For the purposes of planning, the quadcopter's 3D state is given by its position in space $x_R = [p_x, p_y, p_z]$, with velocity controls assumed decoupled in each spatial direction, up to $v_R=0.25$~m/s.
The human can only move by walking and therefore her state is given by planar coordinates $x_H = [h_x, h_y]$ evolving as $\dot{x}_H = [v_H \cos\phi_H, v_H \sin\phi_H]$.
At any given time, the human is assumed to either move at a leisurely walking speed (${v_H \approx 1}$~m/s) or remain still (${v_H \approx 0}$).
\\ \indent
In this example, $\mathcal{K}$ consists of joint robot-human states in which the quadcopter is flying within a square of side length $l=0.3$~m centered around the human's location, while at
any altitude, as well as any joint states in which the robot is outside the bounds of a box with a square base of side $L=3.66$~m and height $H=2$~m, regardless of the human's state.
}

\subsection{Robot Dynamics}

Ideally, robots should plan their motion based on a high-fidelity model of their dynamics, accounting for inertia, actuator limits, and environment disturbances.
Unfortunately, reasoning with such complex models is almost always computationally prohibitive.
 As a result, the models used for planning typically constitute a simplified representation of the physical dynamics of the real robot, and are therefore subject to some error that can have critical implications for safety. In particular, let $s_R\in\mathbb{R}^{n_S}$ denote the state of the robot in the higher-fidelity dynamical model, and let ${\pi:\mathbb{R}^{n_S}\to\mathbb{R}^{n_R}}$ be a known function that projects this higher-fidelity state onto a corresponding planning state, i.e $x_R = \pi(s_R)$. A planner which operates on $x_R$ may generate a trajectory which is difficult to track or even infeasible under the more accurate dynamical model.
Thus reasoning with the planning model alone is not sufficient to guarantee safety for the real robot.

\example{
    We model our quadcopter with the following flight dynamics (in near-hover regime):\\
    \noindent
    \begin{equation}
        \small \label{eq:Quad6D_dyn}
        \begin{bmatrix}
            \dot p_{x}\\ \dot p_{y}\\ \dot p_{z}
        \end{bmatrix}
        = \begin{bmatrix}
            v_{x} \\ v_{y} \\ v_{z} \
        \end{bmatrix}
        \enspace ,\begin{bmatrix}
             \dot v_{x}\\ \dot v_{y}\\ \dot v_{z}\\
        \end{bmatrix}
        = \begin{bmatrix}
              g \tan\theta\\ -g \tan\phi \\ \tau - g 
        \end{bmatrix}
        \enspace,
    \end{equation}
    where $[p_x,p_y,p_z]$ is the quadcopter's position in space and $[v_x,v_y,v_z]$ is its velocity expressed in the fixed world frame, with thrust $\tau$ and attitude angles (roll $\phi$ and pitch $\theta$) as controls.
    The quadcopter's motion planner generates nominal kinematic trajectories in the lower-dimensional $[p_x,p_y,p_z]$ position state space.
    Therefore we have a linear projection map ${\pi(s_R) = [I_3, 0_3] s_R}$, that is, $x_R$ retains the position variables in $s_R$ and discards the velocities.
}

\subsection{Predictive Human Model}

The robot has a predictive model of the human's motion, based on a set of parameters whose values may be inferred under a Bayesian framework or otherwise estimated over time.
Extensive work in econometrics and cognitive science has shown that human behavior can be well modeled by utility-driven optimization \cite{von1945theory, luce1959individual, baker2007goal}.
Thus, the robot models the human as optimizing a reward function, $r_H(x_H,u_H; \theta)$, that depends on the human's state and action, as well as a set of parameters $\theta$. This reward function could be a linear combination of features as in many inverse optimal control implementations (where the weighting $\theta$ between the features needs to be learned), or more generally learned through function approximators such as deep neural networks (where~$\theta$ are the trained weights) \cite{finn2016guided}.

We assume that the robot has a suitable human reward function, either learned offline from prior human demonstrations or otherwise encoded by the system designers. 
With this, the robot can compute the human's policy as a probability distribution over actions conditioned on the state.
Using maximum-entropy assumptions \cite{ziebart2008maximum} and inspiration from noisy-rationality models used in cognitive science \cite{baker2007goal}, the robot models the human as more likely to choose actions with high expected utility, in this case the state-action value (or Q-value):
   \begin{equation}
    \label{eq:observation_model}
        P(u^t_H \mid x^t_H; \beta, \theta) = \frac{e^{\beta Q_H(x^t_H,u^t_H; \theta)}}{\sum_{\tilde u}e^{\beta Q_H(x^t_H,\tilde u; \theta)}}\enspace.
    \end{equation}

\example{
    The quadcopter's model of the human assumes that she intends to reach some target location $g\in\mathbb{R}^2$ in the most direct way possible. The human's reward function is given by the distance traveled $ 
        \label{eq:reward}
        r_H(x_H,u_H;g) = - ||u_H||_2 \enspace$
    and human trajectories are constrained to terminate at $g$.
    The state-action value, parametrized by ${\theta=g}$, captures the optimal cost of reaching $g$ from $x_H$ when initially applying $u_H$:
    $
        \label{eq:value}
        Q_H(x_H,u_H;g) = -||u_H||_2 - ||x_H+u_H-g||_2 .
    $
    }

The coefficient $\beta$ is traditionally called the \textit{rationality coefficient} and it determines the degree to which the robot expects to observe human actions aligned with its model of utility. 
A common interpretation of $\beta = 0$ is a human who appears ``irrational," choosing actions uniformly at random and completely ignoring the modeled utility,    while $\beta \rightarrow \infty$ corresponds a ``perfectly rational" human.
Instead, we believe that $\beta$ can be given a more pragmatic interpretation related to the accuracy with which the robot's model of the human is able to explain her motion.
Consistently, in this paper, we refer to $\beta$ as \textit{model confidence}.

Note that we assume the human does not react to the robot. This assumption can realistically capture plausible shared-space settings in which lightweight robots (e.g. micro-drones) may be expected to carry out services such as indoor surveillance in a building while minimizing interference with human activity.
Additionally, to the extent that a more compliant human will tend to avoid collisions with the robot, the robot may still benefit in such scenarios---it is merely not assuming any cooperation \textit{a priori} in its planning.

\subsection{Probabilistic Safe Motion Planning Problem}
\label{subsec:prob_safe_motion_planning}

The problem that the robot needs to solve is to plan a trajectory that, when tracked by the physical system, will reach a goal state as efficiently as possible while avoiding collisions with high confidence, based on an informed prediction of the human's future motion.

Since any theoretical guarantee is tied to the model it is based on, 
safety guarantees will inherit the probabilistic nature of human predictions.
This induces a fundamental tradeoff between safety and \textit{liveness}:
predictions of human motion may assign non-zero probability to a wide range of states at a future time, which may severely impede the robot's ability to operate in the shared space with ``absolute safety''
(only absolute according to the model).
Therefore, depending on the context, the designers or operators of the system will need to determine what is an acceptable probability that a robot's plan will conflict with the human's future motion.
Based on this, the robot's online planning algorithm will determine when a motion plan is predicted to be \textit{sufficiently safe}. In our demonstrated system, we use a $1\%$ collision probability threshold for planning. 

Our goal now is to find efficient robot motion plans that will keep collisions with a human below an acceptable probability.
Formally, given a current state $x_R^{\text{now}}\in\mathbb{R}^{n_R}$, a cumulative cost ${c:\mathbb{R}^{n_R} \times \mathbb{R}^{m_R} \to \mathbb{R}}$, a probability threshold $P_{\text{th}}\in[0,1]$ and a final time $T$, we define the constrained planning problem:
\begin{subequations}
\begin{align}
        \min_{u_R^{t:T}} &\sum_{\tau=t}^T c(x_R^\tau,u_R^\tau) \label{eq:opt_objective}\\
        \text{s.t. } & x_R^t = x_R^{\text{now}} \label{eq:opt_initial}\\
        &x_R^{\tau+1} = \tilde f_R(x_R^\tau,u_R^\tau), \quad \tau\in{t,...,T-1}\label{eq:opt_dynamics}\\
        & P^{t:T}_\text{coll}:=P\big(\exists \tau\in\{t,...,T\}: \text{coll}(x_R^\tau,x_H^\tau)\big) \le P_{\text{th}}\label{eq:opt_safety}
\end{align}
\end{subequations}
with $\tilde f_R$ a discrete-time approximation of the dynamics~$f_R$. The term $\text{coll}(x_R^t,x_H^t)$ is a Boolean variable indicating whether the human and the robot are in collision.
The safety analysis necessary to solve this online motion planning problem therefore has two main components, the robot's state and the human's state, both of which are affected by uncertainty in their evolution over time.
We tackle these two sources of uncertainty through a combined method that draws simultaneously on the two main approaches to uncertain systems: probabilistic and worst-case analysis.

\example{
The quadcopter's cost can be a weighted combination of distance traversed and time elapsed on its way to a specified goal: $c(x_R,u_R) = ||u_R||_2 + c_0$.
}

The proposed approach in this paper follows two central steps to provide a quantifiable, high-confidence collision avoidance guarantee for the robot's motion around the human. 
In Section \ref{sec:solution_pred} we present our proposed Bayesian framework for reasoning about the uncertainty inherent in a model's prediction of human behavior. 
Based on this inference, we demonstrate how to generate a real-time probabilistic prediction of the human's motion over time. 
Next, in Section \ref{sec:solution_plan} we introduce a theoretical extension to a state-of-the-art, provably safe, real-time robotic motion planner to incorporate our time-varying probabilistic human predictions yielding a quantitative probabilistic safety certificate.

\section{Confidence-Aware Human Motion Prediction}

\label{sec:solution_pred}


Predictions of human motion, even when based on well-informed models, may eventually perform poorly when the human's behavior outstrips the model's predictive power.
Such situations can have a negative impact on safety if the robot fails to appropriately, and quickly, notice the degradation of its predictions.

It will often be the case in practice that the same model will perform variably well over time in different situations and for different people. In some cases this model might be perfectly representative, in others the robot might not have access to some important feature that explains the human's behavior, and therefore the robot's conservativeness should vary accordingly.

Given a utility-based human model in the form of \eqref{eq:observation_model}, the $\beta$ term can be leveraged as an indicator of the model's predictive capabilities, rather than the human's actual level of \textit{rationality}.
Thus, by maintaining an estimate of $\beta$, the robot can dynamically adapt its predictions (and therefore its motion plan) to the current reliability of its human model.
For this reason, in this paper, we refer to $\beta$ as \textit{model confidence},
and aim to make the robot reason about its value in real time in order to generate confidence-aware ``introspective" predictions of the human's motion.

\subsection{Real-time Inference of Model Confidence}
At every time step $t$, the robot obtains a new measurement%
\footnote{In practice, the robot measures the evolution of the human state and computes the associated action by inverting the motion model.}
of the human's action, $u^t_H$. This measurement can be used as evidence to update the robot's belief $b^t(\cdot)$ about $\beta$ over time via a Bayesian update:
    \begin{equation}
    \label{eq:belief_update}
        b^{t+1}(\beta) = \frac{P(u^t_H \mid x^t_H; \beta, \theta)b^{t}(\beta)}{\sum_{\hat\beta} P(u^t_H \mid x^t_H; \hat\beta,\theta)b^{t}(\hat\beta)} \enspace ,
    \end{equation}
    with $b^t(\beta) = P(\beta| x_H^{0:t})$ for $t\in\{0,1,...\}$, and ${P(u^t_H | x^t_H; \beta, \theta)}$ given by \eqref{eq:observation_model}.
It is critical to be able to perform this update extremely fast, which would be difficult to do in the original continuous hypothesis space $\beta\in[0,\infty)$ or even a large discrete set.
Fortunately, as we will see in Section~\ref{sec:demo}, maintaining a Bayesian belief over a relatively small set of $\beta$ values ($N_\beta\approx10$ on a log-scale) achieves significant improvement relative to maintaining a fixed precomputed value.%
\footnote{
Since the predictive performance of the model might change over time as the human's behavior evolves,
we do not in fact treat $\beta$ as a static parameter, but as a hidden state in a hidden Markov model (HMM).
Concretely, between successive ``measurement updates" \eqref{eq:belief_update}, we apply a uniform smoothing ``time update", allowing our belief over $\beta$ to slowly equalize over time, which has the effect of downweighting older observations.
}

\subsection{Human motion prediction}
We can now use the belief over $\beta$ to recursively propagate the human's motion over time and obtain a probabilistic prediction of her state at any number of time steps 
into the future.
In particular, at every future time step, we can estimate the likelihood of the human taking action $u_H$ from any state $x_H$ by directly applying \eqref{eq:observation_model}.
Combining this with the dynamics model, we can generate a distribution of human occupancies over time, with the recursive update:
    \begin{equation}\label{eq:pred_recursive}
    \begin{split}
        P(&x_H^{\tau+1} \mid x_H^{\tau};\beta,\theta) =\\
        &\sum_{u_H^{\tau}}P(x_H^{\tau+1} \mid x_H^{\tau},u_H^{\tau};\beta,\theta)P(u_H^\tau \mid x_H^{\tau};\beta,\theta) \enspace ,
    \end{split}
    \end{equation}
     for $\tau\in\{t,...,T\}$;
     for the deterministic dynamics in our case, $P(x_H^{\tau+1} \mid x_H^\tau, u_H^\tau;\beta, \theta)=\mathds{1}\{x_H^{\tau+1}=\tilde f_H(x_H^\tau,u_H^\tau)\}$. 

\example{
    \begin{figure}
    \includegraphics[width=.48\textwidth]{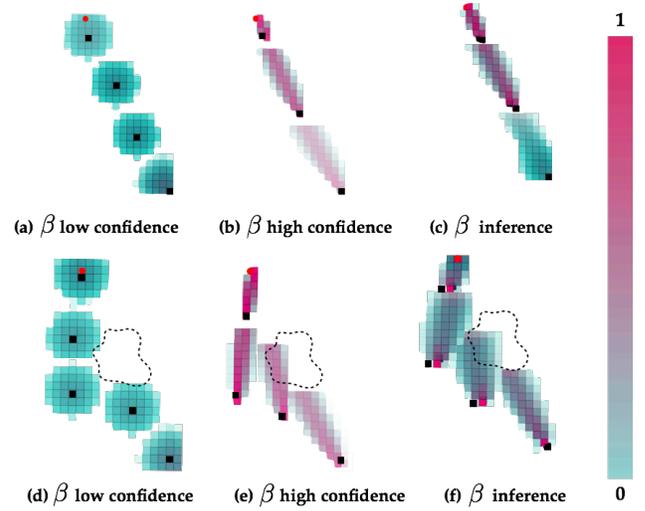} 
    \caption{Snapshots of human trajectory and probabilistic model predictions. Top row: Human moves from the bottom right to a goal marked as a red circle. Bottom row: Human changes course to avoid a spill on the floor. The first two columns show the predictions for low and high model confidence; the third column shows the predictions using our Bayesian model confidence.}
    \vspace{-0.5cm}
    \label{fig:coffee_six}
    \end{figure}
    The simplest scenario in our running example involves a human moving towards a known goal. In Fig \ref{fig:coffee_six}(a-c), the human acts predictably, moving directly to the goal. Each subfigure shows the robot's human prediction under different confidence conditions. Predictions for the second scenario, where the human deviates from her path to avoid a coffee spill on the ground, are shown in Fig \ref{fig:coffee_six}(d-f).
    %
    }


\subsection{Integrating Model Confidence into Online Model Updates}

When a robot is faced with human behavior that is not well explained by its current model, it can attempt to update some of its elements to better fit the observed human actions.
These elements can include parameters, hyperparameters, or potentially even the structure of the model itself. 
Assuming that the parameters can be tractably adjusted online, this update may result in better prediction performance.

Even under online model updates, it continues to be necessary for the robot to reason about model confidence.
In this section we demonstrate how reasoning about model confidence can be done compatibly (and in some cases jointly) with model parameter updates. 


Recall that $\theta$ denotes the set of parameters in the human's utility model. The ideal approach is to perform inference over both the model confidence, $\beta$, and the model parameters, $\theta$ by maintaining a joint Bayesian belief, $b^t(\beta, \theta)$. 
The joint Bayesian belief update rule takes the form
\begin{equation}
    b^{t+1}(\beta,\theta) = \frac{P(u^t_H \mid x^t_H; \beta, \theta)b^t(\beta, \theta)}{\sum_{\hat{\beta}, \hat{\theta}} P(u^t_H \mid x^t_H; \hat{\beta}, \hat{\theta})b^t(\hat{\beta}, \hat{\theta})}
    \enspace ,
\end{equation}
with $b^t(\beta,\theta) = P(\beta,\theta \mid x^{0:t}_H, u^{0:t}_H)$.%
\footnote{
Analogously to the case with $\beta$-only inference, the parameters $\theta$ can be allowed to evolve as a hidden state.
}
This approach can be practical for parameters taking finitely many values from a discrete set, for example, possible distinct modes for a human driver (distracted, cautious, aggressive).

\begin{figure}
\includegraphics[width=.48\textwidth]{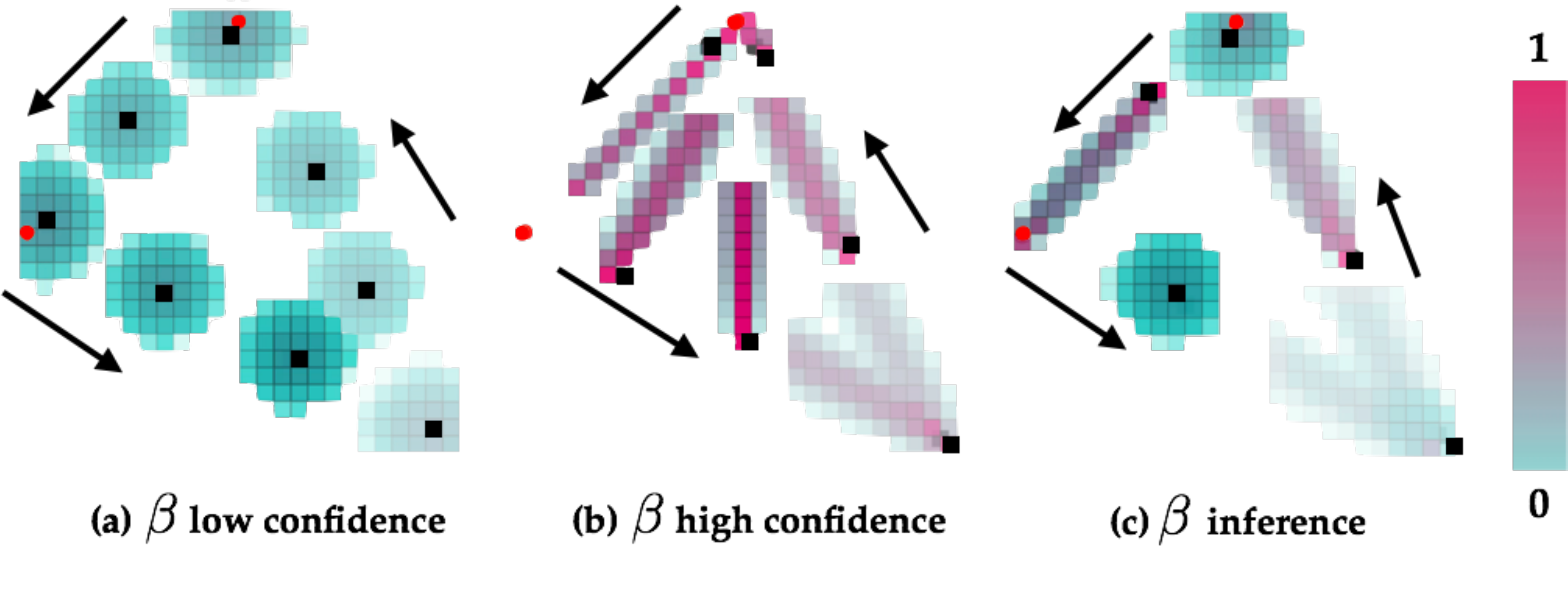} 
\caption{In this example the human is moving in a counter-clockwise motion to two goals (marked in red), and then to a third unknown goal (located at the same position as the start). Subfigures (a) and (b) show the predictions for a low and high $\beta$, respectively.
Subfigure (c) shows the predictions using our inferred model confidence, where the robot is confident when the human is moving ``rationally'', and uncertain when the human behavior does not match the robot's model.}
\label{fig:triangle_three}
\vspace{-0.7cm}
\end{figure}

\example{
The quadcopter's model of the human considers a number of known frequently-visited locations $\theta\in\{g_1,...,g_N\}$ that she might intend to walk to next.
However, there may be additional unmodeled destinations, or more complex objectives driving the human's motion in the room (for example, she could be searching for a misplaced object, or pacing while on the phone). Fig. \ref{fig:triangle_three} shows how reasoning about model confidence \textit{as well as} the human's destination enables the robot to navigate confidently while the human's motion is well explained by the model, and automatically become more cautious when it departs from its predictions. More detailed results are presented in Section~\ref{sec:demo}.
}

For certain scenarios or approaches it may not be practical to maintain a full Bayesian belief on the parameters, and these are instead estimated over time (for example, through a maximum likelihood estimator (MLE), or by shallow re-training of a pre-trained neural network). In these cases, a practical approach can be to maintain a ``bootstrapped" belief on $\beta$ by running the Bayesian update on the running parameter estimate $\bar\theta$:
    \begin{equation}
        \label{eq:bootstrap}
        \bar{b}^{t+1}(\beta) = \frac{P(u^t_H \mid x^t_H; \beta, \bar{\theta})\bar{b}^{t}(\beta)}{\sum_{\hat\beta} P(u^t_H \mid x^t_H; \hat\beta,\bar{\theta})\bar{b}^{t}(\hat\beta)}
        \enspace .
    \end{equation}
\example{
The quadcopter's predictions of human motion are parameterized by her walking speed $v_H$;
the quadcopter maintains a simple running average based on recent motion-capture measurements,
and incorporates the current estimate into inference and prediction.
}

When it is not desirable or computationally feasible to update the parameter estimate $\bar{\theta}$ continually, we can leverage our model confidence as an indicator of when re-estimating these parameters may be most useful---namely as confidence in the model under the current parameter estimates degrades.


\section{Safe Probabilistic Planning and Tracking} \label{sec:solution_plan}

Once it can generate real-time probabilistic predictions of the human's motion, the robot needs to plan a trajectory that will, with high probability, avoid collisions with her.
On the one hand, any rigorous safety analysis for a robotic system needs to account for deviations of the actual dynamic trajectory from the ideal motion plan.
On the other hand, since human motion predictions are by nature uncertain, the safety analysis will necessarily be quantified in probabilistic terms.
To this end, we build on the recent FaSTrack framework \cite{herbert2017fastrack}, which provides control-theoretic robust safety certificates in the presence of deterministic obstacles, and extend the theoretical analysis to provide probabilistic certificates allowing uncertain dynamic obstacles (here, humans). 

\subsection{Background: Fast Planning, Safe Tracking}
\label{subsec:fastrack}
Recall that $x_R$ and $u_R$ are the robot's state and control input, for the purposes of motion planning.
The recently proposed FaSTrack framework \cite{herbert2017fastrack} uses Hamilton-Jacobi reachability analysis~\cite{Evans1984,Mitchell2005} to provide a simple real-time motion planner with a worst-case tracking error bound and error feedback controller for the dynamic robot.
Formally, FaSTrack precomputes an optimal tracking control policy, as well as a corresponding compact set $\mathcal{E}$ in the robot's planning state space, such that ${\big(\pi (s_R^t) - x_{R,\text{ref}}^t\big)\in \mathcal{E}}$ for \textit{any} reference trajectory proposed by the lower-fidelity planner.
This bound $\mathcal{E}$ is a trajectory tracking certificate that can be passed to the online planning algorithm for real-time safety verification:
the dynamical robot is guaranteed to \textit{always} be somewhere within the bound relative to the plan.
Therefore the planner can generate safe plans by ensuring that the entire bound around the nominal state remains collision-free throughout the trajectory.
Note that the planner only needs to know $\mathcal{E}$ and otherwise requires no explicit understanding of the high-fidelity model.

\example{
    Since dynamics \eqref{eq:Quad6D_dyn} are decoupled in the three spatial directions,
    the bound $\mathcal{E}$
    computed by FaSTrack
    is an axis-aligned box of dimensions  $\mathcal{E}_x \times \mathcal{E}_y\times \mathcal{E}_z$.
}

\subsection{Robust Tracking, Probabilistic Safety}

Unfortunately, planning algorithms for collision checking against deterministic obstacles cannot be readily applied to our problem.
Instead, a trajectory's collision check should return the probability that it \textit{might} lead to a collision.
Based on this probability, the planning algorithm can discriminate between trajectories that are \textit{sufficiently safe} and those that are not.

As discussed in Section~\ref{subsec:prob_safe_motion_planning}, a safe online motion planner should continually check
the probability that, at any future time $\tau$,
${(\pi (s_R^\tau),x_H^\tau)\in\mathcal{K}}$.
The tracking error bound guarantee from FaSTrack allows us to conduct worst-case analysis on collisions given a human state $x_H$:
if no point in the Minkowski sum $\{x_R+\mathcal{E}\}$ is in the collision set with $x_H$, we can guarantee that the robot is not in collision with the human.

The probability of a collision event for any point $x_R^\tau$ in a candidate trajectory plan, assuming worst-case tracking error, can be computed as the total probability that $x_H^\tau$ will be in collision with \textit{any} of the possible robot states $\tilde{x}_R\in\{x_R^\tau+\mathcal{E}\}$.
For each robot planning state $x_R\in\mathbb{R}^{n_R}$ we can define the set of human states in potential collision with the robot:
\begin{equation}\label{eq:H_E}
   \mathcal{H}_\mathcal{E}(x_R) := \{ \tilde{x}_H\in\mathbb{R}^{n_H}: \exists \tilde{x}_R\in\{x_R+\mathcal{E}\}, (\tilde{x}_R,\tilde{x}_H)\in\mathcal{K} \}
   \enspace .
\end{equation}
The following result is then true by construction.

\textbf{{Proposition 1:}}\emph{
    The probability of a robot with worst-case tracking error $\mathcal{E}$ being in collision with the human at any trajectory point $x_R^\tau$ is bounded above by the probability mass of $x_H^\tau$ contained within $\mathcal{H}_\mathcal{E}(x_R^\tau)$.
    }

Therefore,
the left-hand side of the inequality in our problem's safety constraint \eqref{eq:opt_safety} can be rewritten as 
\begin{equation}
    P^{t:T}_\text{coll} = 1-\prod_{\tau=t}^T P\big(x_H^\tau\not\in \mathcal{H}_\mathcal{E}(x_R^\tau)
    \mid x_H^\tau\not\in \mathcal{H}_\mathcal{E}(x_R^s), t\le s<\tau\big) 
    .
\end{equation}

Evaluating the above probability exactly would require reasoning jointly about the distribution of human states over all time steps, or equivalently over all time trajectories $x_H^{0:T}$ that the human might follow.
Due to the need to plan in real time, we must in practice approximate this distribution.

Since assuming independence of collision probabilities over time is both unrealistic and overly conservative, we instead seek to find a tight lower bound on a trajectory's overall collision probability based on the marginal probabilities at each moment in time. In particular, based on the positive correlation
over time resulting from human motion continuity,
we first consider replacing each conditional probability
$P\big(x_H^\tau\not\in \mathcal{H}_\mathcal{E}(x_R^\tau)
    \mid x_H^s\not\in \mathcal{H}_\mathcal{E}(x_R^s), t\le s<\tau\big)$
    by $1$ for all $t>0$.
This would then give the lower bound
\begin{equation}\label{eq:lower_bound}
    P^{t:T}_\text{coll}\ge 1-P\big(x_H^t\not\in \mathcal{H}_\mathcal{E}(x_R^t)\big)
    = P\big(x_H^t\in \mathcal{H}_\mathcal{E}(x_R^t)\big)
    \enspace ,
\end{equation}
which would seem like an unreasonably optimistic approximation.
However, note that probabilities can be conditioned in any particular order (not necessarily chronological) and we can therefore generate $T-t+1$ lower bounds of the form $P^{t:T}_\text{coll}\ge P\big(x_H^\tau\in \mathcal{H}_\mathcal{E}(x_R^\tau)\big)$ 
for $\tau\in\{t,\hdots,T\}$, again by replacing all successive conditional non-collision probabilities by~1.
Taking the tightest of all of these bounds, we can obtain an informative, yet quickly computable, approximator for the sought probability:
\begin{equation}\label{eq:prob_approx}
    P^{t:T}_\text{coll} \approx \max_{\tau\in\{t:T\}}P\big(x_H^{\tau}\in \mathcal{H}_\mathcal{E}(x_R^{\tau})\big)
    \enspace .
\end{equation}
In other words, we are replacing the probability of collision of an entire trajectory with the highest marginal collision probability at \textit{each point} in the trajectory.
While this approximation will err on the side of optimism, we note that the robot's ability to continually replan as updated human predictions become available mitigates any potentially underestimated risks, since in reality the robot does not need to commit to a plan that was initially deemed safe, and will readily rectify as the estimated collision risk increases prior to an actual collision.

\example{
Given $\mathcal{K}$ and $\mathcal{E}$,
$\mathcal{H}_\mathcal{E}(x_R^\tau)$ is the set of human positions within the rectangle of dimensions ${(l+\mathcal{E}_x) \times (l+\mathcal{E}_y)}$ centered on $[p_x^\tau,p_y^\tau]$.
A human anywhere in this rectangle could be in collision with the quadcopter.
}

\subsection{Safe Online Planning under Uncertain Human Predictions}

We can now use this real-time evaluation of collision probabilities to discriminate between valid and invalid trajectory candidates in the robot's online motion planning.
Using the formulation in Section~\ref{sec:solution_pred}, we can quickly generate, at every time~$t$, the marginal probabilities in \eqref{eq:prob_approx} at each future time $\tau \in\{t,\hdots,T\}$, based on past observations at times $0,\hdots,t$.
Specifically, for any candidate trajectory point $x_R^\tau$, we first calculate the set $\mathcal{H}_\mathcal{E}(x_R^\tau)$;
this set can often be obtained analytically from \eqref{eq:H_E}, and can otherwise be numerically approximated from a discretization of $\mathcal{E}$.
The planner then computes the instantaneous probability of collision $P\big(x_H^\tau\in \mathcal{H}_\mathcal{E}(x_R^\tau)\big)$ by integrating $P\big(x_H^{\tau} \mid x_H^{0:t}\big)$ over $\mathcal{H}_\mathcal{E}(x_R^\tau)$, and rejects the candidate point $x_R^\tau$ if this probability exceeds $P_{\text{th}}$.

Note that for search-based planners that consider candidate trajectories by generating a tree of timestamped states, rejecting a candidate node from this tree is equivalent to rejecting all further trajectories that would contain the node. This early rejection rule is consistent with the proposed approximation \eqref{eq:prob_approx} of $P^{t:T}_\text{coll}$ while preventing unnecessary exploration of candidate trajecories that would ultimately be deemed unsafe.


As the robot is continuously regenerating its motion plan online as the human's predicted motion is updated, we simultaneously track the planned trajectory using our error feedback controller, which ensures that we deviate by no more than the tracking error bound $\mathcal{E}$. This planning and tracking procedure continues until the robot's goal has been achieved.
\\ \indent
\begin{figure}
\includegraphics[width=.48\textwidth]{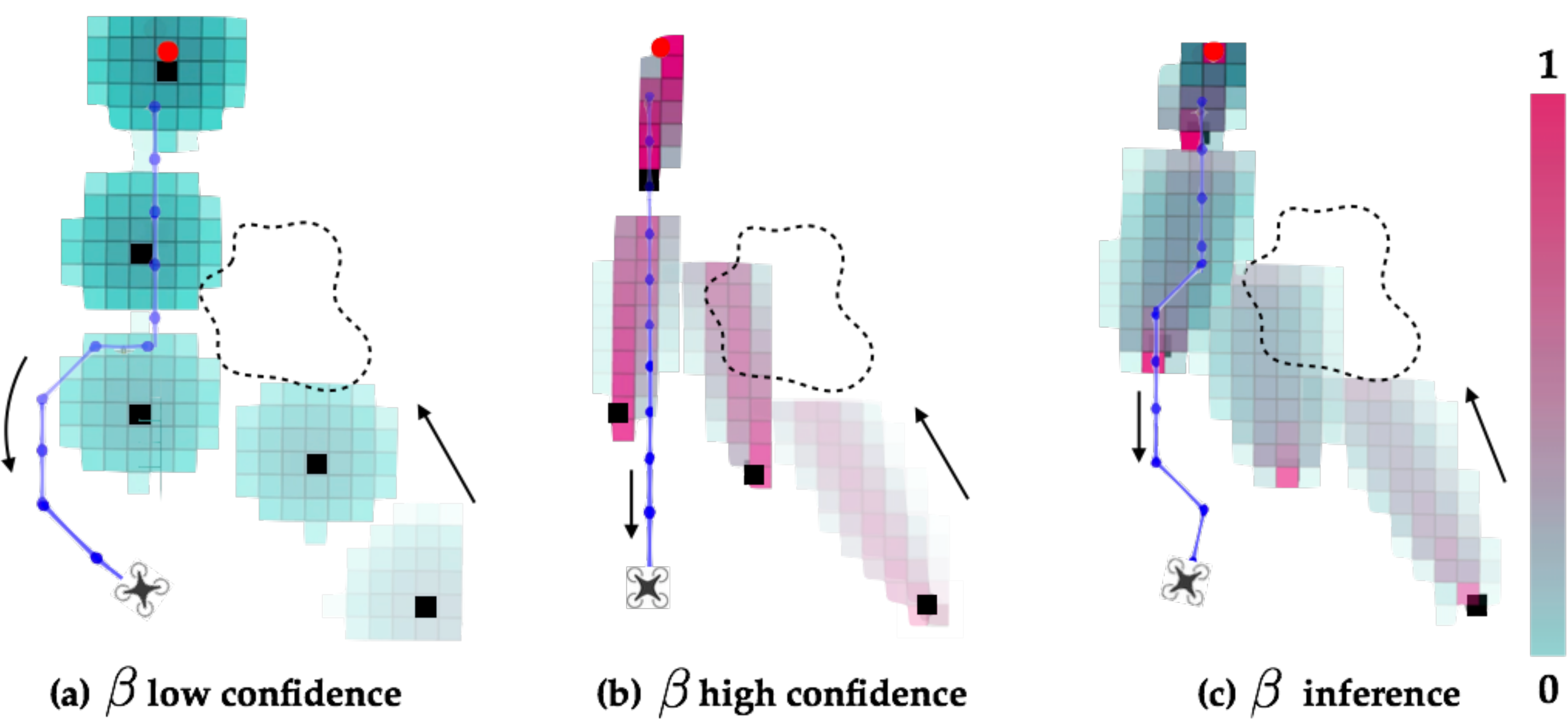} 
\caption{Scenario from Fig. \ref{fig:coffee_six} visualized with robot's trajectory based on its current $\beta$. When $\beta$ is low and the robot is not confident, it makes large deviations from its path to accommodate the human. When $\beta$ is high, the robot refuses to change course and comes dangerously close to the human. With inferred model confidence, the robot balances safety and efficiency with a slight deviation around the human.}
\label{fig:coffee_three_quadrotor}
\vspace{-0.7cm}
\end{figure}
\example{
    Our quadcopter is now required to navigate to a target position shown in Fig. \ref{fig:coffee_six} without colliding with the human. Our proposed algorithm successfully avoids collisions at all times, replanning to leave greater separation from the human whenever her motion departs from the model. In contrast, robot planning with fixed model confidence is either overly conservative at the expense of time and performance or overly aggressive at the expense of safety.
}

\section{Demonstration with Real Human Trajectories}
\label{sec:demo}

We implemented real-time human motion prediction with $\beta$ inference and safe probabilistic motion planning via FaSTrack within the Robot Operating System (ROS) framework \cite{quigley2009ros}. To demonstrate the characteristic behavior of our approach, we created three different environment setups and collected a total of 48 human walking trajectories (walked by 16 different people). The trajectories are measured as $(x, y)$ positions on the ground plane at roughly 235 Hz by an OptiTrack infrared motion capture system.\footnote{We note that in a more realistic setting, we would need to utilize alternative methods for state estimation such as lidar measurements.} 
We also demonstrated our system in hardware on a Crazyflie 2.0 platform navigating around a person in a physical space.\footnote{\url{https://youtu.be/2ZRGxWknENg}}  

\textbf{Environments.} In the first environment there are no obstacles and the robot is aware of the human's goal. The second environment is identical to the first, except that the human must avoid a coffee spill that the robot is unaware of. In the third environment, the human walks in a triangular pattern from her start position to two known goals and back. 


\textbf{Evaluated Methods.} For each human trajectory, we compare the performance of our adaptive $\beta$ inference method with two baselines using fixed $\beta \in \{0.05, 10\}$. When $\beta = 0.05$, the robot is unsure of its model of the human's motion. This low-confidence method cannot trust its own predictions about the human's future trajectory. On the other hand, the $\beta = 10$ high-confidence method remains confident in its predictions even when the human deviates from them. These two baselines exist at opposite ends of a spectrum. Comparing our adaptive inference method to these baselines provides useful intuition for the relative performance of all three methods in common failure modes (see Fig.~\ref{fig:coffee_three_quadrotor}).

\textbf{Metrics.} We measure the performance of our adaptive $\beta$ inference approach in both of these cases by simulating a quadcopter moving through the environment to a pre-specified goal position while replaying the recorded human trajectory. We simulate near-hover quadcopter dynamics with the FaSTrack optimal controller applied at 100 Hz. For each simulation, we record the minimum distance in the ground plane between the human and the quadcopter as a proxy for the overall safety of the system. The quadcopter's travel time serves to measure its overall efficiency.

In each environment, we compute the safety metric for all 16 human trajectories when applying each of the three human motion prediction methods and display the corresponding box and whisker plots side by side. To compare the efficiency of our approach to the baselines we compute the difference between the trajectory completion time of our approach, $T_{\textnormal{infer}}$, and that of the low and high confidence baselines, $\{T_{\textnormal{lo}}, T_{\textnormal{hi}}\}$. If the resulting boxplots are below zero, then $\beta$ inference results in faster robot trajectories than the baselines on a per-human trajectory basis.\footnote{The upper and lower bounds of the box in each boxplot are the $75^{\text{th}}$ and $25^{\text{th}}$ percentiles. The horizontal red line is the median, and the notches show the bootstrapped $95\%$ confidence interval for the population mean.}


\begin{figure*}[htbp]
    \centering
    \includegraphics[width=0.95\textwidth]{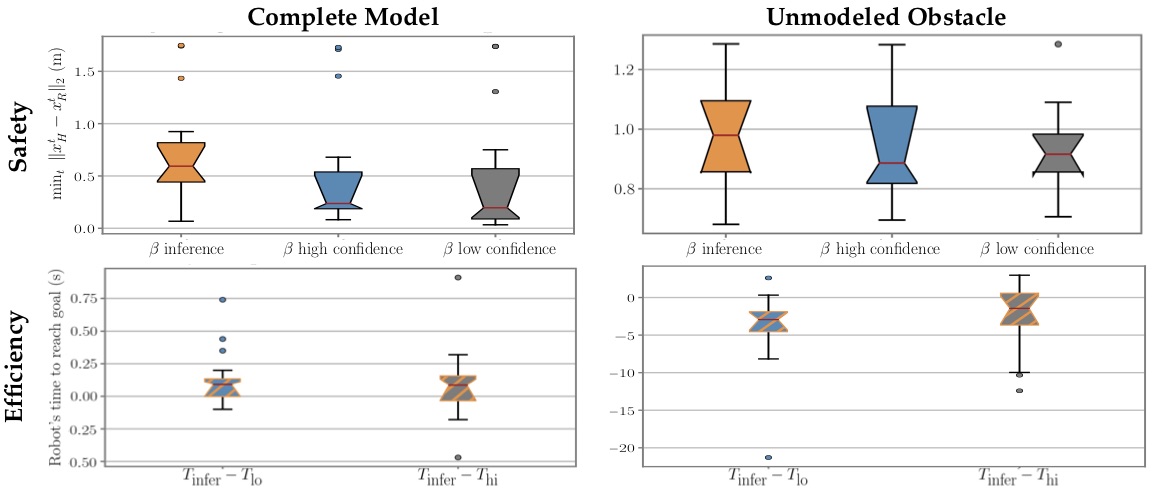}
    \caption{Safety and efficiency metrics in a complete environment and one with an unmodeled obstacle.}
    \label{fig:coffee_nocoffee}
    \vspace{-.55cm}
\end{figure*}

\textbf{Complete Model.}
\label{subsec:complete_model} First, we designed an example environment where the robot's model is complete and the human motion appears to be rational. In this scenario, humans would walk in a straight line from their start location to their goal which was known by the robot \textit{a priori}. 

When the robot has high confidence in its model, the human's direct motion towards the goal appears highly rational and results in both safe (Fig.~\ref{fig:coffee_nocoffee}, top left) and efficient plans (Fig.~\ref{fig:coffee_nocoffee}, bottom left). We see a similar behavior for the robot that adapts its confidence: although initially the robot is uncertain about how well the human's motion matches its model, the direct behavior of the human leads to the robot to believe that it has high model confidence. Thus, the $\beta$ inference robot produces overall safe and efficient plans. Although we expect that the low-confidence model would lead to less efficient plans but comparably safe plans, we see that the low-confidence robot performs comparably in terms of both safety and efficiency. 

Ultimately, this example demonstrates that when the robot's model is rich enough to capture the environment and behavior of the human, inferring model confidence does not hinder the robot from producing safe and efficient plans.

\textbf{Unmodeled Obstacle.}
\label{subsec:unmodeled_obstacle} Often, robots do not have fully specified models of the environment. In this scenario, the human has the same start and goal as in the complete model case except that there is a coffee spill in her path. This coffee spill on the ground is unmodeled by the robot, making the human's motion appear less rational. 

When the human is navigating around the unmodeled coffee spill, the robot that continuously updates its model confidence and replans with the updated predictions almost always maintains a safe distance (Fig.~\ref{fig:coffee_nocoffee}, top right). In comparison, the fixed-$\beta$ models that have either high-confidence or low-confidence approach the human more closely. This increase in the minimum distance between the human and the robot during execution time indicates that continuous $\beta$ inference can lead to safer robot plans. 

For the efficiency metric, a robot that uses $\beta$ inference is able to get to the goal faster than a robot that assumes a high or a low confidence in its human model (Fig.~\ref{fig:coffee_nocoffee}, bottom right). This is particularly interesting as overall we see that enabling the robot to reason about its model confidence can lead to \textit{safer} and \textit{more efficient} plans. 

\textbf{Unmodeled Goal.}
\label{subsec:unmodeled_goal} In most realistic human-robot encounters, even if the robot does have an accurate environment map and observes all obstacles, it is unlikely for it to be aware of all human goals. We test our approach's resilience to unknown human goals by constructing a scenario in which the human moves between both known and unknown goals. The human first moves to two known goal positions, then back to the start. The first two legs of this trajectory are consistent with the robot's model of goal-oriented motion. However, when the human returns to the start, she appears irrational to the robot.
\begin{figure}[htbp]
    \centering
    \includegraphics[width=\columnwidth]{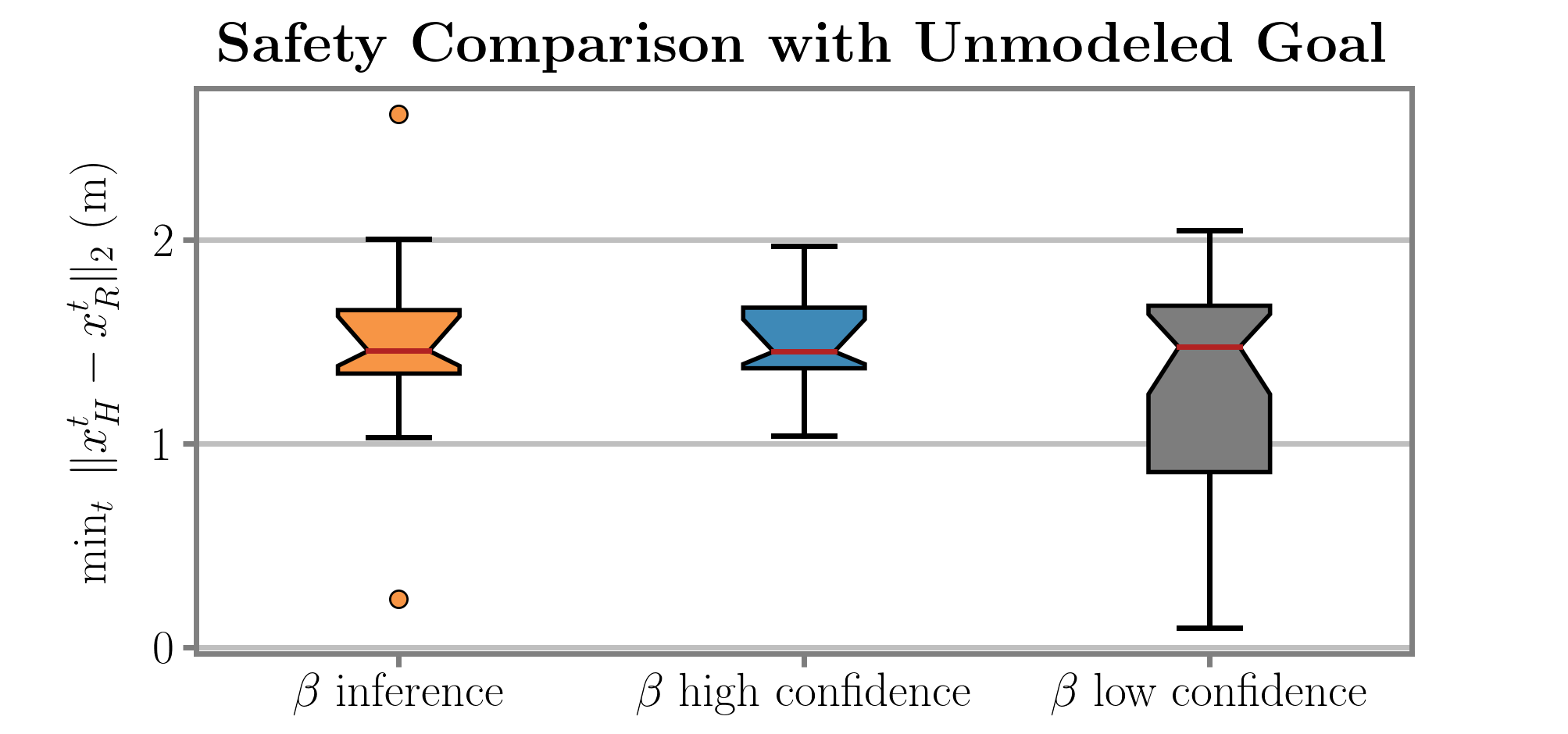}
    \caption{Safety results for the unmodeled goal scenario.}
    \label{fig:safe_triangle_results}
\end{figure}
\begin{figure}[htbp]
    \centering
    \includegraphics[width=\columnwidth]{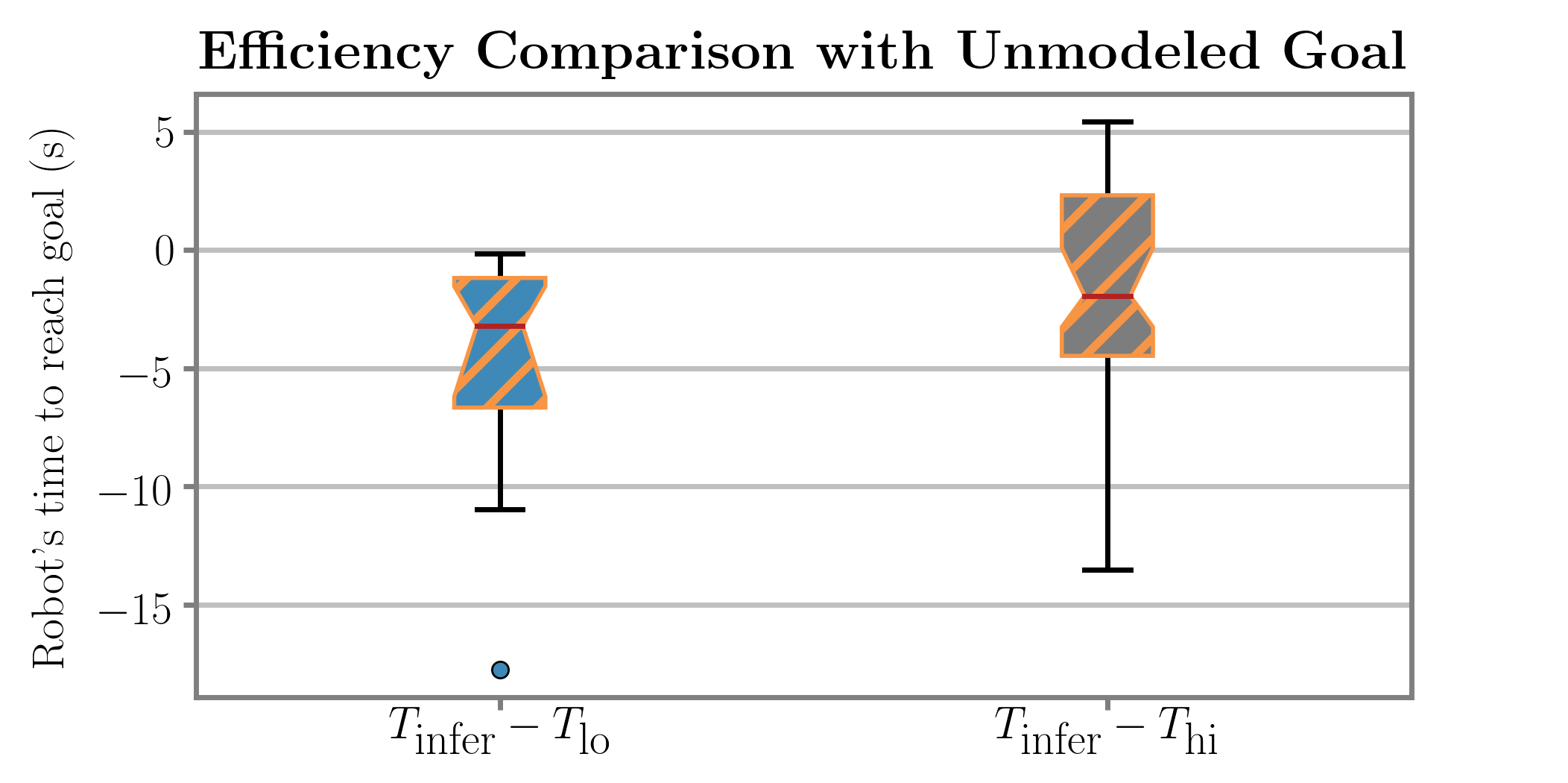}
    \caption{Efficiency results for the unmodeled goal scenario.}
    \label{fig:eff_triangle_results}
    \vspace{-0.5cm}
    
\end{figure}
Fig.~\ref{fig:safe_triangle_results} and \ref{fig:eff_triangle_results} summarize the performance of the inferred-$\beta$, high-confidence, and low-confidence methods in this scenario. All three methods perform similarly with respect to the minimum distance safety metric in Fig.~\ref{fig:safe_triangle_results}. However, Fig.~\ref{fig:eff_triangle_results} suggests that the inferred-$\beta$ method is several seconds faster than both fixed-$\beta$ approaches. This indicates that, without sacrificing safety, our inferred-$\beta$ approach allows the safe motion planner to find more efficient robot trajectories.

\vspace{-.1cm}
\section{Discussion \& Conclusion}
\label{sec:conclusion}
\vspace{-.1cm}
In this paper, we interpret the ``rationality'' coefficient in the human decision modeling literature as an indicator of the robot's confidence in its ability to predict human motion. We formulate this confidence $\beta$ as a hidden state that the robot can infer by contrasting observed human motion with its predictive model. 
Marginalizing over this hidden state, the robot can quickly adapt its forecasts to effectively reflect the predictability of the human's motion in real time. We build on the theoretical analysis of the provably safe FaSTrack motion planning scheme to construct a novel probabilistic safety certificate that combines worst-case and probabilistic analysis, and show that the resulting trajectories are collision-free at run-time with high probability. 

We compare our $\beta$ inference technique to two fixed-$\beta$ approaches, all using our proposed probabilistically safe motion planning scheme. Our results indicate that, even though the three methods perform similarly when the human's motion is well-explained by the robot's model, inferring $\beta$ yields safer and more efficient robot trajectories in environments with unmodeled obstacles or unmodeled human goals. 
Future work should investigate more complex human motion, closed-loop interaction models, and navigating around multiple humans.






\section*{Acknowledgments}
We thank Smitha Milli for confidence inference guidance.

\bibliographystyle{plainnat}
\bibliography{references}

\end{document}